\documentclass{article}
\usepackage{test}

\usepackage[margin=1in]{geometry}

\usepackage{graphicx}
\usepackage{booktabs,arydshln}
\makeatletter
\def\adl@drawiv#1#2#3{%
        \hskip.5\tabcolsep
        \xleaders#3{#2.5\@tempdimb #1{1}#2.5\@tempdimb}%
                #2\z@ plus1fil minus1fil\relax
        \hskip.5\tabcolsep}
\newcommand{\cdashlinelr}[1]{%
  \noalign{\vskip\aboverulesep
          \global\let\@dashdrawstore\adl@draw
          \global\let\adl@draw\adl@drawiv}
  \cdashline{#1}
  \noalign{\global\let\adl@draw\@dashdrawstore
          \vskip\belowrulesep}}
\makeatother

\newcommand{\gain}[1]{\colorbox[HTML]{d3ff9f}{+#1}}

\newcommand{\highlightpink}[1]{\colorbox[HTML]{fbbad7}{\textbf{#1}}}

\newcommand{\highlightblue}[1]{\colorbox[HTML]{bae6fb}{\textbf{#1}}}

\usepackage{array}
\usepackage{amsmath}
\usepackage{amssymb}
\usepackage{amsfonts}
\usepackage{multirow}
\usepackage{verbatim}
\usepackage{caption}
\usepackage{longtable}
\usepackage{supertabular}
\usepackage{float}

\usepackage{algorithm,algpseudocode}

\usepackage{enumitem}
\usepackage{tablefootnote}
\usepackage[round,semicolon]{natbib}
\usepackage{xcolor}
\usepackage{colortbl}
\usepackage{xspace}
\usepackage{textcomp}
\usepackage{makecell}
\usepackage{multirow}
\usepackage{lscape} 
\usepackage{siunitx}

\usepackage{amssymb}%
\usepackage{pifont}%
\usepackage{scrextend}

\usepackage{array}
\usepackage{tgpagella}
\usepackage{latexsym}
\usepackage[T1]{fontenc}
\usepackage[utf8]{inputenc}
\usepackage{microtype}
\definecolor{mydarkblue}{rgb}{0,0.08,0.45}
\usepackage[colorlinks,citecolor=mydarkblue,urlcolor=mydarkblue,linkcolor=mydarkblue]{hyperref}
\usepackage{url}            %
\usepackage{nicefrac}       %
\usepackage{changepage}
\usepackage{xargs}          %

\usepackage{wrapfig,lipsum,booktabs}
\usepackage{longtable}
\usepackage{subcaption}
\usepackage{endnotes}

\usepackage{pgfplots}
\usetikzlibrary{pgfplots.groupplots}
\pgfplotsset{compat=1.3}
\usepackage{tikz}
\usetikzlibrary{patterns}

\usepackage[most]{tcolorbox}

\usepackage[capitalize,noabbrev]{cleveref}
\crefname{section}{Section}{\S\S}
\Crefname{section}{Section}{\S\S}
\crefname{table}{Table}{Tables}
\crefname{figure}{Figure}{Figures}
\crefname{algorithm}{Algorithm}{}
\crefname{equation}{eq.}{}
\crefname{appendix}{Appendix}{}
\crefformat{section}{Section #2#1#3}
\usepackage{multicol}

\usepackage{todonotes}
\usepackage{titlesec}
\titleformat*{\section}{\large\bfseries}

\newcommand{\Meta}{Meta-prompting\xspace}
\newcommand{\meta}{meta-prompting\xspace}

\definecolor{battleshipgrey}{rgb}{0.3, 0.3, 0.3}
\definecolor{brilliantrose}{rgb}{1.0, 0.33, 0.64}
\definecolor{americanrose}{rgb}{1.0, 0.01, 0.24}
\definecolor{jweigreen}{rgb}{0,0.45,0.24}
\definecolor{bluegray}{rgb}{0.1, 0.1, 0.4}
\definecolor{ao(english)}{rgb}{0.0, 0.5, 0.0}
\definecolor{blanchedalmond}{rgb}{1.0, 0.92, 0.8}
\definecolor{atomictangerine}{rgb}{1.0, 0.6, 0.4}
\definecolor{chocolate(web)}{rgb}{0.82, 0.41, 0.12}
\definecolor{bananayellow}{rgb}{1.0, 0.88, 0.21}
\definecolor{goldenbrown}{rgb}{0.6, 0.4, 0.08}
\definecolor{aliceblue}{rgb}{0.94, 0.97, 1.0}
\definecolor{beige}{rgb}{0.96, 0.96, 0.86}
\definecolor{babyblue}{rgb}{0.54, 0.81, 0.94}
\definecolor{camel}{rgb}{0.76, 0.6, 0.42}
\definecolor{cinnamon}{rgb}{0.82, 0.41, 0.12}
\definecolor{deepskyblue}{rgb}{0.0, 0.75, 1.0}
\definecolor{frenchblue}{rgb}{0.0, 0.45, 0.73}
\definecolor{classicrose}{rgb}{0.98, 0.8, 0.91}
\definecolor{frenchrose}{rgb}{0.96, 0.29, 0.54}
\definecolor{frenchlilac}{rgb}{0.53, 0.38, 0.56}
\definecolor{frenchbeige}{rgb}{0.65, 0.48, 0.36}
\definecolor{applegreen}{rgb}{0.55, 0.71, 0.0}
\definecolor{dartmouthgreen}{rgb}{0.05, 0.5, 0.06}
\definecolor{columbiablue}{rgb}{0.61, 0.87, 1.0}
\definecolor{rufous}{rgb}{0.66, 0.11, 0.03}
\definecolor{cyan(process)}{rgb}{0.0, 0.72, 0.92}
\definecolor{crimsonglory}{rgb}{0.75, 0.0, 0.2}

\usepackage{tcolorbox}

\usepackage[font=small,labelfont=bf]{caption}
\usepackage[capitalize]{cleveref}

\title{\textbf{Meta-Prompting}:\\\vspace{0.2em} \large{Enhancing Language Models with Task-Agnostic Scaffolding}}

\author{
\normalsize{\textbf{Mirac Suzgun}} \\
\normalsize{Stanford University}\thanks{Work done while at Microsoft Research New England.} \\
\normalsize{\url{msuzgun@stanford.edu}}
\and
\normalsize{\textbf{Adam Tauman Kalai}}\\
\normalsize{OpenAI}$^*$\\
\normalsize{\url{adam@kal.ai}}
}

\date{}

\begin{document}

\maketitle

\vspace{-1em}
\begin{abstract}
\noindent
We introduce \meta, an effective scaffolding technique designed to enhance the functionality of language models (LMs). This approach transforms a single LM into a multi-faceted conductor, adept at managing and integrating multiple independent LM queries. By employing high-level instructions, \meta guides the LM to break down complex tasks into smaller, more manageable subtasks. These subtasks are then handled by distinct ``expert'' instances of the same LM, each operating under specific, tailored instructions. Central to this process is the LM itself, in its role as the conductor, which ensures seamless communication and effective integration of the outputs from these expert models. It additionally employs its inherent critical thinking and robust verification processes to refine and authenticate the end result. This collaborative prompting approach empowers a single LM to simultaneously act as a comprehensive orchestrator and a panel of diverse experts, significantly enhancing its performance across a wide array of tasks. The zero-shot, task-agnostic nature of \meta greatly simplifies user interaction by obviating the need for detailed, task-specific instructions. Furthermore, our research demonstrates the seamless integration of external tools, such as a Python interpreter, into the \meta framework, thereby broadening its applicability and utility. Through rigorous experimentation with GPT-4, we establish the superiority of \meta over conventional scaffolding methods: When averaged across all tasks, including the Game of 24, Checkmate-in-One, and Python Programming Puzzles, \meta---augmented with a Python interpreter functionality---surpasses standard prompting by 17.1\%, expert (dynamic) prompting by 17.3\%, and multipersona prompting by 15.2\%.\footnote{
The data, prompts, and the model outputs are all available at \url{https://github.com/suzgunmirac/meta-prompting}.}
\end{abstract}

\vspace{-0.8em}
\begin{figure*}[hb]
\begin{centering}
\begin{tikzpicture}
  \pgfplotsset{footnotesize,samples=10}
    \begin{groupplot}[
        group style = {group size = 4 by 1, horizontal sep = 30pt},
        width = 6.0cm, 
        height = 5.2cm,
                    cycle list={
                {fill=crimsonglory}, %
                {fill=frenchblue}, %
            },
]
        \nextgroupplot[
            align = center,
            title = {\textbf{Game of 24}},
            legend style={
                at={(0.4,1.0)},
                anchor=north,
                legend columns=1,
                nodes={anchor=west} %
            },
            xmin=0.6, xmax=6.4,
            ymin=0, ymax=103,
            xtick={1,2,3,4,5,6},
            xticklabels={Std, 0-CoT, Ex-St, Ex-Dy, MP, \textbf{Meta}},
            xticklabel style = {font=\scriptsize},
            axis x line*=bottom,
            axis y line*=left,
            ylabel={Task Accuracy (\%)},
            ytick={0, 20, 40, 60, 80, 100},
            yticklabels={0, 20, 40, 60, 80, 100},
            grid style=dashed,
            x label style={at={(axis description cs:0.5,-0.12)},anchor=north},
            y label style={at={(axis description cs:-0.12,0.5)},anchor=south},
            xtick pos=bottom,
            ytick pos=left,
            ]

            \addplot+[
                ybar, bar width=10pt,             
                ]
                coordinates {
                (1, 3)
                (2, 11)
                (3, 3)
                (4, 2)
                (5, 25)
                }; \addplot+[ ybar, bar width=10pt, ] coordinates { %

                (6, 67)
                };

        \nextgroupplot[
            align = center,
            title = {\textbf{Checkmate-in-One}},
            xmin=0.6, xmax=6.4,
            ymin=0, ymax=103,
            xtick={1,2,3,4,5,6},
            xticklabels={Std, 0-CoT, Ex-St, Ex-Dy, MP, \textbf{Meta}},
            xticklabel style = {font=\scriptsize},
            axis x line*=bottom,
            axis y line*=left,
            ytick={0, 20, 40, 60, 80, 100},
            grid style=dashed,
            x label style={at={(axis description cs:0.5,-0.12)},anchor=north},
            y label style={at={(axis description cs:-0.12,0.5)},anchor=south},
            xtick pos=bottom,
            ytick pos=left,
            ]

            \addplot+[
                ybar, bar width=10pt,                ]
                coordinates {
                (1, 36.4)
                (2, 32.8)
                (3, 39.6)
                (4, 33.2)
                (5, 17.2)
                                }; \addplot+[ ybar, bar width=10pt, ] coordinates { %
                (6, 57.2)
             };

\nextgroupplot[
    align = center,
    title = {\textbf{Sonnet Writing}},
    xmin=0.6, xmax=6.4,
            ymin=0, ymax=103,
            xtick={1,2,3,4,5,6},
            xticklabels={Std, 0-CoT, Ex-St, Ex-Dy, MP, \textbf{Meta}},
            xticklabel style = {font=\scriptsize},
            axis x line*=bottom,
            axis y line*=left,
            ytick={0, 20, 40, 60, 80, 100},
            grid style=dashed,
            x label style={at={(axis description cs:0.5,-0.12)},anchor=north},
            y label style={at={(axis description cs:-0.12,0.5)},anchor=south},
            xtick pos=bottom,
            ytick pos=left,
    ]

      \addplot[
                ybar, fill=crimsonglory, bar width=10pt,                ]
                coordinates {
        (1, 62.0)
        (2, 71.2)
        (3, 74)
        (4, 74)
        (5, 73.2)
                        }; \addplot+[ ybar, bar width=10pt, ] coordinates { %
        (6, 79.6)
        };
    \end{groupplot}
\end{tikzpicture}
\caption{
Enhancing GPT-4 with \meta. In this study, we introduce and examine the effectiveness of meta-prompting, contrasting it with a range of zero-shot prompting techniques, including 
standard zero-shot (Std), 
zero-shot chain-of-thought (0-CoT;~\citet{ZeroShotCoT}), 
generic and dynamic expert (Ex-St and Ex-Dy;~\citet{xu2023expertprompting}), 
and multipersona (MP;~\citet{wang2023soloprompting}).
Our research demonstrates that meta-prompting, particularly when combined with a Python interpreter, significantly improves overall accuracy and robustness in GPT-4 across a variety of tasks.
} 
\label{fig:scale-cot-interaction}
\end{centering}
\end{figure*}
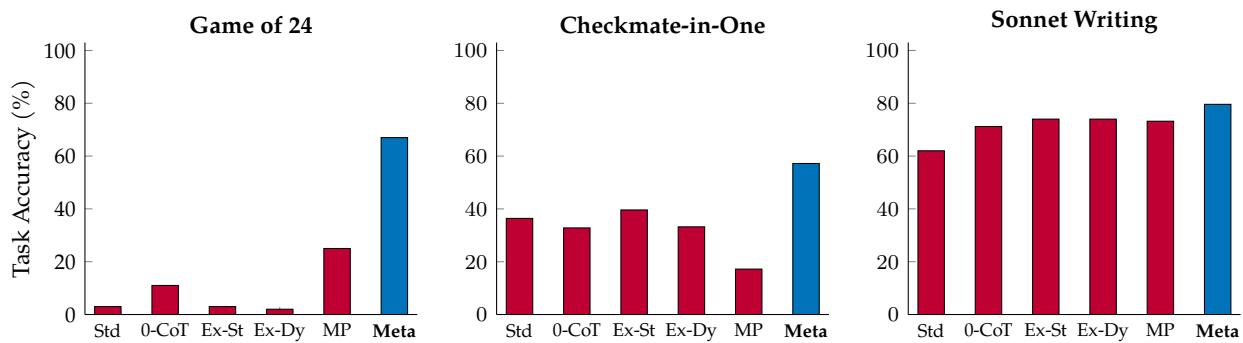

\clearpage

\section{Introduction}
The latest generation of language models (LMs)---notably, GPT-4~\citep{GPT4}, PaLM~\citep{PaLM2}, and LLaMa~\citep{touvron2023llama}---have expanded the boundaries of natural-language processing and generation. These large-scale models can tackle a wide spectrum of tasks, ranging from writing Shakespearean sonnets about hedgehogs to summarizing intricate medical reports and solving competition-level programming puzzles. Despite their versatility, these models are not infallible; they sometimes generate responses that are inaccurate, misleading, or conflicting. As the operational costs of these models become more affordable, it becomes natural to ask whether one might use scaffolding systems and leverage multiple LM queries to not only refine but also to enhance the accuracy and robustness of these model outputs.

In this work, we introduce a new technique for enhancing the functionality and performance of LMs, called \meta. It involves constructing a high-level ``meta'' prompt that instructs an LM to: (i) break down complex tasks or problems into smaller, manageable pieces; (ii) assign these pieces to specialized ``expert'' models with proper and detailed natural-language instructions; (iii) oversee the communication between these expert models; and (iv) apply its own critical thinking, reasoning, and verification skills throughout the process. When presented with a query, the LM, effectively prompted under \meta, serves as a conductor. It produces a message history---a narrative, if you will---comprising the responses from various expert models. The LM is originally responsible for generating the conductor's portion of this history, which includes the selection of experts and the formulation of specific instructions for them. However, the same LM doubles itself as these independent experts as well, generating outputs based on the expertise and information chosen by the conductor for each particular query. 

This approach allows for a single, uniform LM to maintain a coherent line of reasoning while also tapping into a variety of expert roles. The use of dynamically selected contexts for prompting these experts introduces fresh perspectives into the process, while the conductor model retains a bird's-eye view of the entire history and coordination. This method, therefore, enables a single black-box LM to function effectively as both a central conductor and a diverse panel of experts to produce more accurate, reliable, and coherent responses.

Our proposed \meta~technique combines and expands upon various prompting ideas introduced by recent studies---including, 
\emph{high-level planning and decision-making}~\citep{yao2023react,sun2023adaplanner,hao2023reasoning},
\emph{dynamic persona assignment}~\citep{xu2023expertprompting,wang2023soloprompting},
\emph{multi-agent debating}~\citep{du2023multiagent,zhuge2023mindstorms},
\emph{self-debugging and self-reflection}~\citep{schick2023peer,liu2023reflect,gou2023critic,madaan2023selfrefine,shinn2023reflexion}.
A key aspect of \meta is its \emph{task-agnostic} nature. Unlike traditional scaffolding methods that require specific instructions or examples tailored to each task, \meta employs the same set of high-level instructions across various tasks and inputs. This universality is particularly beneficial for users who might find it cumbersome to provide detailed examples or specific guidance for every distinct task. For instance, in responding to a one-off request like ``Write a Shakespearean sonnet about selfies,'' the user would not need to supply examples of high-quality neoclassical poems. The \meta approach elevates the utility of language models by offering a broad, flexible framework without compromising on specificity or relevance. Additionally, to demonstrate the versatility and integration capabilities of \meta, we have enhanced our system with the functionality to invoke a Python interpreter. This allows for an even more dynamic and comprehensive application of the technique, further extending its potential to address a wide array of tasks and queries effectively.

\begin{figure*}[t]
    \centering
    \includegraphics[width=0.99\linewidth]{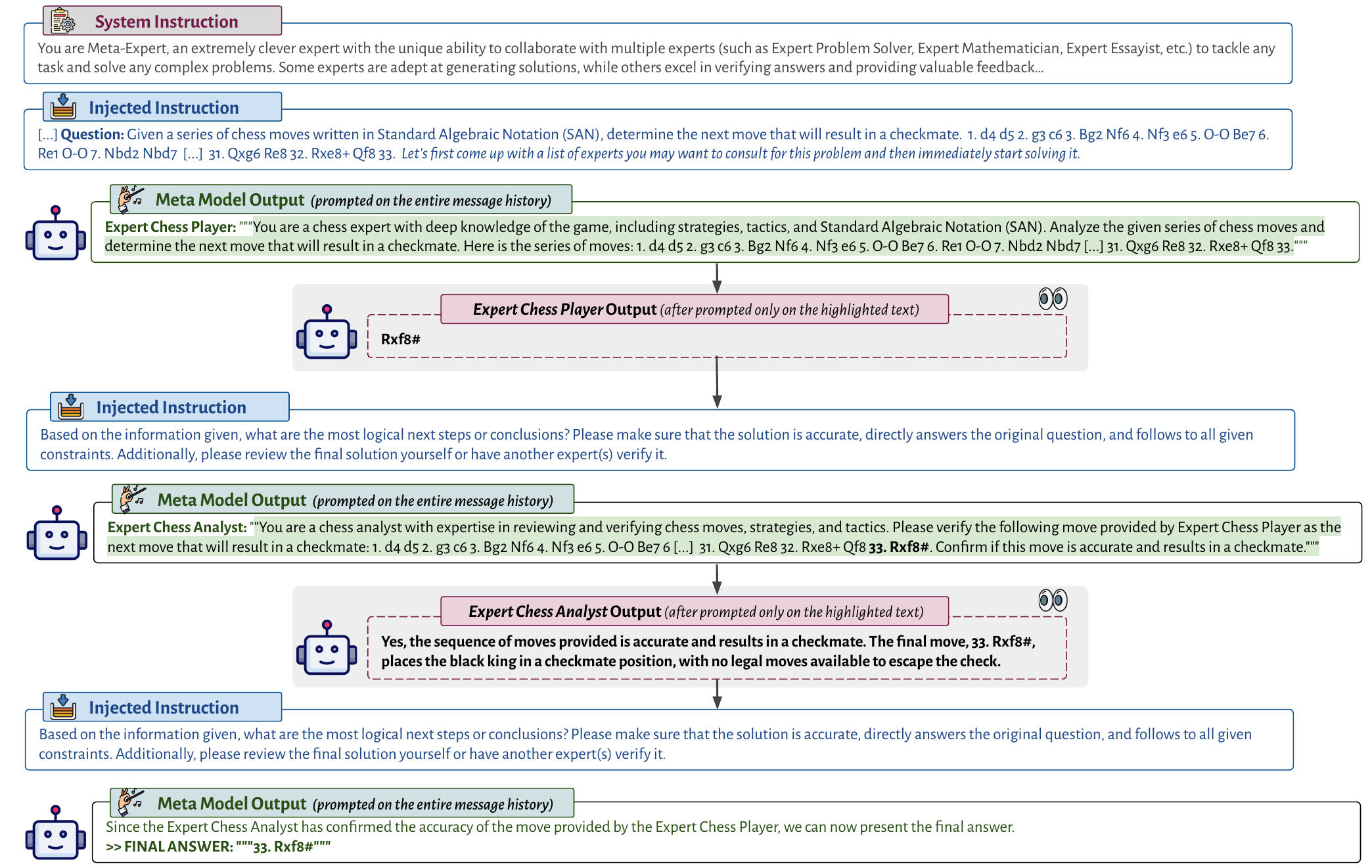}
    \caption{An example \meta history, where the prompts have been shortened for illustrative purposes. The history is initialized by a question provided by a user. Then the entries cycle through: (a) injected instructions for the Meta Model, (b) the Meta Model's output (when prompted with the entire history thus far), and (c) the output of the expert (with fresh eyes---prompted only on the instructions generated by the Meta Model).
    }
    \label{fig:example}
\end{figure*}

We provide an illustrative visualization of a \meta session in Figure~\ref{fig:example}. It depicts how the Meta Model---our technical term for the central controlling LM (a.k.a. the conductor)---intersperses its own output with inputs and outputs from various specialized expert models or code executions. Such a configuration makes \meta a nearly universal tool. It allows for the consolidation of various LM interactions and computations into a single, coherent narrative. What sets \meta apart is that it leaves the decision of which prompts to use and which code snippets to execute to the discretion of the LM itself.

In our comprehensive experiments, which primarily utilize GPT-4 as the foundational LM, we compare the efficacy of \meta against other task-agnostic scaffolding methods. Our findings reveal that \meta not only enhances overall performance but often leads to state-of-the-art results across a diverse range of tasks. Its flexibility is noteworthy: The conductor model has the capability to call upon expert models (basically itself, albeit with fresh instructions) for performing a variety of functions. These functions might include critiquing earlier outputs, selecting specific personas for certain tasks, refining generated content, and ensuring that the final outputs meet the desired criteria in both substance and form. This approach shows a marked improvement over several existing methods, as demonstrated in~Figure~\ref{fig:scale-cot-interaction}.

The core contribution of this work is the introduction of a task-agnostic scaffolding system that leverages a single LM. This LM not only carries forward the thread of the task but also dynamically selects and instructs expert models appropriate for each specific task. The effectiveness of this system is showcased across various benchmarks, including the Game of 24~\citep{yao2023tree}, Checkmate-in-One from the BIG-Bench suite~\citep{srivastava2023beyond}, and our novel task of  ``Shakespearean Sonnet Writing.''  Overall, our empirical results underscore the versatility and robustness of \meta in enhancing LM performance.

\section{Meta Prompting}
\label{sec:meta-promting}

\textbf{Intuition and Abstract Overview.} 
The modus operandi of \meta is to use a model\footnote{Our use of the term model refers to the application of an LM with certain prompt templates to play a specified ``role.'' We typically only use a single LM (e.g., GPT-4) to implement all the models in an execution.} to coordinate and execute multiple independent inquiries and subsequently synthesize their responses to render a final response. This mechanism, in principle, endorses an ensemble approach, drawing from the strength and diversity of independent specialized models to collaboratively address and tackle multifaceted tasks or problems. We posit that while a single, general-purpose model might  deliver valuable and useful insights into generic queries, combining the perspectives and conclusions of multiple domain-specific models (which we also refer to as \textit{experts}%
) has the potential to yield more comprehensive, robust, and accurate solutions.

Central to our \meta strategy is its shallow hierarchical configuration, where a single model---called the ``Meta Model''---emerges as the principal entity of authority. This prompting structure is reminiscent of an orchestra, wherein the conductor's role is mirrored by the Meta Model and each musician corresponds to a distinct domain-specific model. Just as a conductor harmonizes multiple musical elements to craft a beautiful melody, the Meta Model combines solutions and insights from a range of models to provide an accurate and comprehensive answer to an intricate problem or task.  

Conceptually, a domain-specific expert within our framework can take diverse forms, such as a finetuned LM tailored to perform a particular task, a specialized API equipped to handle specific domain-related inquiries, or even computational tools like calculators or a Python interpreter that can perform arithmetic calculations or write and execute code. These experts, despite their varying functionalities, are directed and unified under the supervision of the Meta Model. 

Under our setup, experts can be called only by the Meta Model. They cannot directly interact or communicate with each other, though the Meta Model can choose to share some text from or combine the insights of various experts when interacting with a new expert. This restriction is made to simplify the communication between the experts and to put the Meta Model at the center of the operation.%

\textbf{Notation and Terminology.}
Before we delve into the specific steps involved in \meta, we establish some notation and terminology. We let $\mathbb{S}$ denote the set of finite strings, with $\emptyset$ representing the empty string. We use $x \in \mathbb{S}$ to refer to a test-time query, which can be a task or a problem described in natural language. A crucial element of \meta is the fixed language model, denoted as $\mathtt{LM}$, which operates from $\mathbb{S}$ to $\mathbb{S}$. This model, like GPT-4, takes an input text (a prompt history that may include a list of previous messages, symbolized by $\mathcal{H}$) and produces a corresponding output (i.e., response).
We also introduce specific template functions: $t_\text{init}, t_\text{mid},$ and $t_\text{exp}$, each mapping from $\mathbb{S}$ to $\mathbb{S}$; each takes a string input and formats it according to a predefined template. Specifically, $t_\text{init}$ and $t_\text{mid}$ are used to format text for the history given to the Meta Model, while $t_\text{exp}$ wraps the output of the Meta Model in a prompt suitable for an expert model.
Furthermore, we have two string extractors, $e_\text{exp}$ and $e_\text{ret}$, each mapping from $\mathbb{S}$ to $\mathbb{S}$. These extractors are designed to retrieve a substring that is enclosed within specific delimiters, returning the first matching segment in cases where multiple segments are present.
The symbol $\oplus$ is used to represent string concatenation. Lastly, we introduce a specific string referred to as $\text{error} \in \mathbb{S}$, which is designed to denote an error message in the process.

\textbf{Algorithmic Procedure.} Algorithm~\ref{alg:main-alg} provides pseudocode of our proposed \meta approach. We further provide a conceptual overview of the procedure below:

\begin{algorithm}
\caption{Meta Prompting}
\textbf{Input}: $\texttt{LM}: \mathbb{S} \rightarrow \mathbb{S}; x, \text{error} \in \mathbb{S}; T \in \mathbb{N}; t_\text{init}, t_\text{mid}, t_\text{exp},  e_\text{exp},e_\text{ret}:\mathbb{S} \rightarrow \mathbb{S}$
\begin{algorithmic}[1] %
\State $\mathcal{H}_{1} \gets t_{\text{init}} (x)$ 
\For{$t \in [1, \ldots, T]$}
    \State $y_{t} \gets \mathtt{LM}~(\mathcal{H}_{t})$
    \If{$e_{\text{exp}} (y_t) \neq \emptyset $} \Comment{Meta Model provided expert instructions}
        \State $\text{prompt} \gets  t_\text{exp}( e_{\text{exp}} (y_t))$
        \State $z_{t} \gets \mathtt{LM}~(\text{prompt})$
        \State $\mathcal{H}_{t+1} \gets \mathcal{H}_{t} \oplus t_{\text{mid}} (z_t)$
    \ElsIf{$e_{\text{ret}} (y_t) \neq \emptyset $} \Comment{Meta Model returned a final answer}
        \State \Return {$e_{\text{ret}} (y_t)$}
    \Else \Comment{Meta Model formatting error}
        \State $\mathcal{H}_{t+1} \gets \mathcal{H}_{t} \oplus \text{error}$
    \EndIf
\EndFor
\end{algorithmic}
\label{alg:main-alg}
\end{algorithm}

\begin{enumerate}

\item \textbf{Transforming the Input}: Using the transformation function $t_{\text{init}}$, the raw query is placed in a suitable template followed by initial instructions to the Meta Model.

\item \textbf{Loop Iteration}:
    \begin{enumerate}
    
    \item \textbf{Prompting the Meta Model}: The current message list, namely $\mathcal{H}_t$, guides the Meta Model's next action---either directly addressing the query or consulting a domain-specific expert.
    
    \item \textbf{Engaging Domain-Specific Expert Models}: If the Meta Model does not return a result, it can conjure any expert and give it instructions, which are extracted from its output using $e_\text{exp}$. This process is isolated though: Each expert only sees what the Meta Model chooses to share with them, and responds accordingly. For instance, if a problem pertains to mathematics and history, the Meta Model might consult a mathematics expert for a calculation and a history expert for historical context. The output of the expert is extracted and additional instructions are appended, all using the $t_\text{mid}$ template.   
    \item \textbf{Returning the Final Response}: If the Meta Model's response contains a final answer (highlighted by distinct special markers), the solution is extracted using $e_\text{ret}$ and returned.
    \item \textbf{Error Handling}: In cases where the model response $y_t$ contains neither a final answer nor a call to an expert model, an error message appended to the message list $\mathcal{H}_t$. This ensures that our procedure is robust and can handle unexpected outputs.
    \end{enumerate}
\end{enumerate}

\textbf{Meta and Expert Model Specifications.} In our setup, we employ the same LM, such as GPT-4, to function in both Meta and Expert capacities. Their roles are distinguished by their respective model instructions in their prompts, with the Meta Model adhering to a set of instructions provided in Figure~\ref{fig:meta-model-instruction}, and the expert models following separate instructions dynamically determined by the Meta Model at inference time .

\section{Experimental Setup}

\subsection{Baselines}
We compare \meta with the task-agnostic, zero-shot versions of the following prompting methods:

\begin{itemize}
    \item \textbf{Standard prompting}: This represents our most basic baseline wherein an LM is asked to directly yield a response without any specific guiding input-output exemplars or any additional guiding instructions, besides the task description already included in the input query.
    \item \textbf{Zero-shot CoT prompting}~\citep{ZeroShotCoT}:
    Drawing inspirations from the chain-of-thought method of \citet{wei2022chain}, this zero-shot prompting approach simply appends ``Let's think step by step'' to the input query, encouraging the model to have a more deliberative and iterative cognition before addressing the problem or task at hand. 
    \item \textbf{Expert prompting}~\citep{xu2023expertprompting}: This prompting approach functions through a two-step process: It first crafts an expert identity tailored to align with the specific context of the input query. It then integrates this generated expert profile into the input to generate a well-informed and authoritative response. In our experiments, we consider two versions of expert prompting, namely (a) \emph{static} (i.e., \emph{generic}) and (b) \emph{dynamic} (i.e., \emph{adaptive}); the former uses a fixed and generic expert description, whereas the latter adaptively designs a new expert identity for each input query.
    \item \textbf{Multi-persona prompting}~\citep{du2023multiagent}: Also known as solo-performance prompting (SPP), this method instructs an LM to perform the following: (i) Propose a small ensemble of ``personas'' to address the specific task or problem at hand; (ii) let these personas engage in a collective dialogue, collaboratively generating potential solutions while extending feedback to one another and refining their answers; and (iii) synthesize all the available information and deliver a final response.
\end{itemize}

\subsection{Datasets and Tasks}
To evaluate the efficacy of our proposed \meta approach over other zero-shot prompting baselines, we consider a wide range of tasks and datasets that require various degrees of mathematical and algorithmic reasoning, domain-specific knowledge, and literary creativity. These include: 
\begin{itemize}
    \item (a) The \textbf{Game of 24} from \citep{yao2023tree} where the goal is to form an arithmetic expression whose value is 24 using each of four given numbers exactly once,
    \item Three BIG-Bench Hard (BBH; \citet{suzgun2023bbh}) tasks---namely, (b) \textbf{Geometric Shapes}, (c) \textbf{Multi-Step Arithmetic Two}, and (d) \textbf{Word Sorting}---as well as one reasoning task directly obtained from the BIG-Bench suite~\citep{srivastava2023beyond}, that is, (e) \textbf{Checkmate-in-One};
    \item (f) \textbf{Python Programming Puzzles} ({P3};~\citet{schuster2021programming}), a collection of challenging programming puzzles written in Python---with varying difficulty levels;
    \item (g) \textbf{Multilingual Grade School Math} ({MGSM}; \citet{shi2023language}), a multilingual version of the GSM8K dataset~\citep{cobbe2021training} with translations of a subset of examples into ten typologically diverse languages, including Bengali, Japanese, and Swahili;
    \item (h) \textbf{Shakespearean Sonnet Writing}, a novel task we created where the goal is to write a sonnet with strict rhyme scheme ``{ABAB CDCD EFEF GG},'' containing the three provided words verbatim.\footnote{While all the other tasks and datasets were previously introduced by other studies, we present this task for the first time.}
\end{itemize}

\subsection{Answer Extraction and Evaluation Protocols}
\begin{figure}[!h]
\centering
\vspace{-0.3em}
\includegraphics[width=0.95\linewidth]{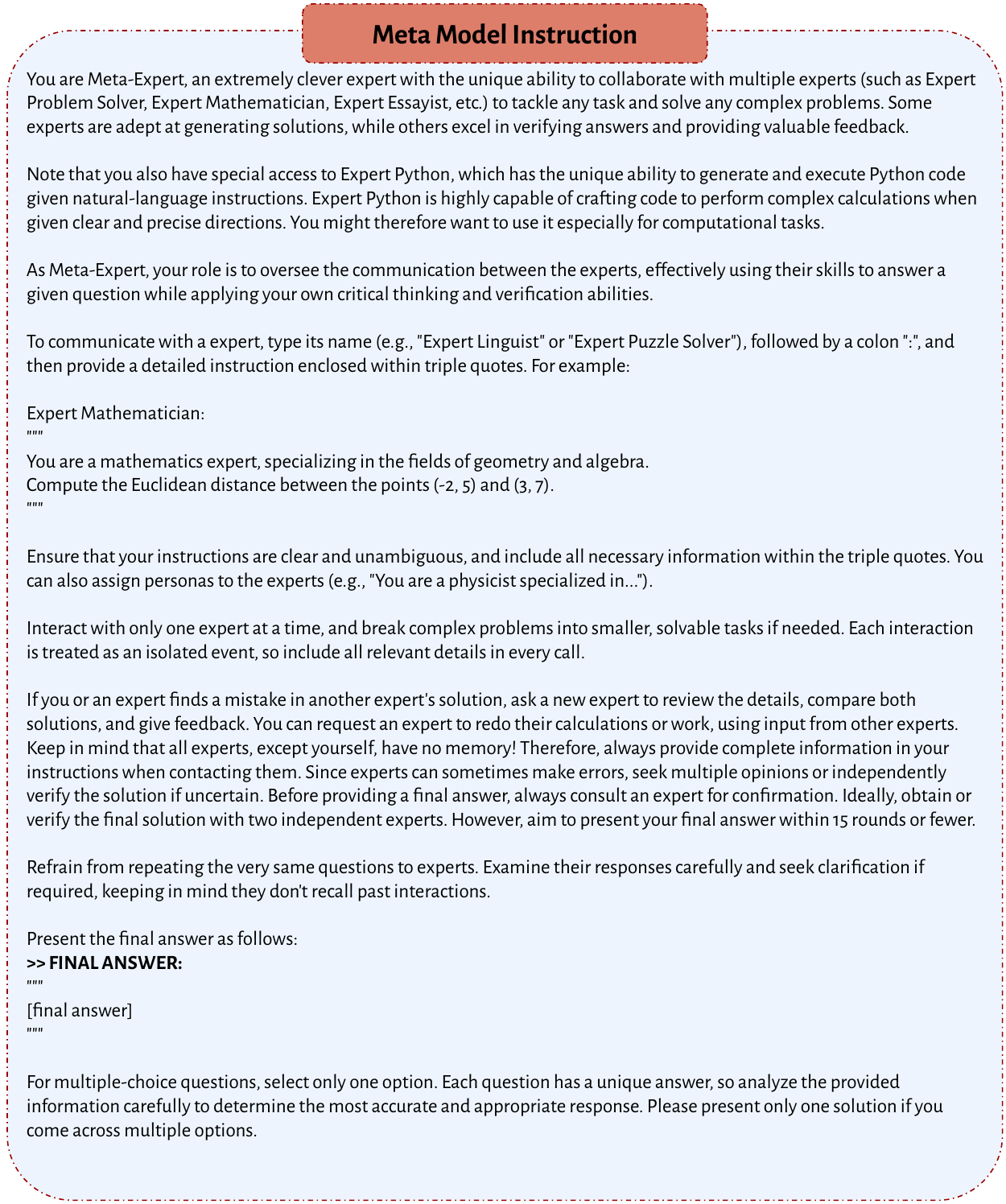}
\vspace{-0.3em}
\caption{The instructions given to the Meta Model using the ``system message'' parameter in the GPT-4 API.}
\label{fig:meta-model-instruction}
\end{figure}

As shown in Figure~\ref{fig:meta-model-instruction}, the system instruction in our proposed \meta method encourages the Meta Model to present its final answer in a specific format. This format, designed for consistent and unambiguous extraction, requires that the final answer is wrapped within triple quotes and preceded by a distinct marker (namely, ``\texttt{>>FINAL ANSWER:}'').

Once the final answer is extracted from the model and properly post-processed, we also need to evaluate its correctness.\footnote{We have developed suitable pipelines for answer extraction and processing tailored to each task. Specific implementation details can be found in our codebase.} Because we consider a wide range of tasks, there is not a single metric that allows us to measure accuracy across all. Depending on the nature and formulation of the task, we measure accuracy using one of the following three metrics:

\begin{itemize}
    \item \textit{Exact Match} (EM): Under this strict metric, the correctness of an answer is determined by its precise alignment with the ground-truth label(s). An answer is deemed correct only if it is identical to a provided reference.
    \item \textit{Soft Match} (SM): This metric offers a more lenient approach than EM. For an answer to be deemed correct, it is sufficient for a ground-truth label to be present within the model's output, regardless of any additional textual content.
    \item \textit{Functionally Correct} (FC): This metric ascertains whether the answer is functionally correct, meaning that it adheres to task-specific constraints.
\end{itemize}

We use EM for Geometric Shapes, Multi-Step Arithmetic Two,  and Checkmate-in-One; SM for MGSM and Word Sorting,; and FC for Game of 24, Python Programming Puzzles, and Shakespearean Sonnet Writing.

\subsection{Models and Inference}
In our main experiments, we concentrate on \textbf{GPT-4} (\texttt{gpt-4-32k}), which is accessible through Microsoft’s Azure OpenAI Service. Additionally, in our supplementary experiments, we include \textbf{GPT-3.5} (\texttt{gpt-35-turbo}). Both GPT-3.5 and GPT-4 are models fine-tuned for following instructions, though GPT-4 has demonstrated significantly better reasoning and content generation abilities than GPT-3.5.\footnote{In our preliminary experiments, we also tested other OpenAI models such as text-davinci-003 and code-davinci-002, but we discovered that our \meta approach yielded consequential results when applied to GPT-3.5 and GPT-4.}

In all of our experiments, we consistently applied the same parameters and system instructions to the Meta Model. We set the temperature value at $0$, the top-p value at $0.95$, and the maximum token count at $1024$.\footnote{The temperature value, which usually ranges between 0 and 1, controls how much randomness or creativity the model exhibits. Ideally, a temperature of 0 should lead to the model producing the same output when presented with the same input. However, both GPT-3.5 and GPT-4 have shown a tendency to generate varied responses even at this setting. This means that reproducing our exact results might be challenging under identical experimental conditions. To address this issue, we are releasing all model inputs, interactions, and outputs in our GitHub repository.} 

\clearpage
\section{Main Results and Discussion}

\begin{table}[!t]
\centering
\scalebox{0.90}{
\begin{tabular}{l|ccccccc|c}
\toprule
& \multicolumn{2}{c}{Basic} & \multicolumn{2}{c}{Expert} & SPP & \multicolumn{2}{c}{Meta} & \multicolumn{1}{|c}{$\Delta$} \\
\cdashlinelr{2-9}
\textbf{Task} &  Standard & 0-CoT & Static & Dynamic & Multi-Persona & \textbf{-} Python & \textbf{+} Python & (M-S) \\
\midrule
{Checkmate-in-One} & \highlightpink{36.4} & 32.8 & 39.6 & 33.2 & 17.2 & \highlightblue{57.2} & \highlightblue{57.2} & \gain{20.8} \\[0.05cm]
{Game of 24} & \highlightpink{3.0} & 11.0 & 3.0 & 2.0 & 25.0 & 11.0 & \highlightblue{67.0} & \gain{64.0} \\[0.05cm]
{Geometric Shapes} & \highlightpink{56.8} & \highlightblue{69.2} & 55.2 & 53.6 & 57.6 & 58.4 & 59.2 & \gain{2.4} \\[0.05cm]
{MGSM (avg)} & \highlightpink{84.4}  & 85.5  & 83.0  & 85.0  & \highlightblue{85.7}  &  85.4 & 84.8  & \gain{0.4} \\[0.05cm]
{Multi-Step Arithmetic} & \highlightpink{84.0} & 83.2 & 83.2 & 78.8 & \highlightblue{91.6} & 84.8 & 90.0 & \gain{6.0} \\[0.05cm]
{Python Prog. Puzzles} & \highlightpink{31.1} & 36.3 & 33.8 & 25.0 & 32.5 & 32.7 & \highlightblue{45.8} & \gain{14.7}  \\[0.05cm]
{Sonnet Writing} & \highlightpink{62.0} & 71.2 & 74.0 & 74.0 & 73.2 & 77.6 & \highlightblue{79.6} & \gain{17.6} \\[0.05cm]
{Word Sorting} & \highlightpink{80.4} & 83.6 & 83.2 & 85.2 & 79.2 & 84.0 & \highlightblue{99.6} & \gain{19.2} \\[0.05cm]
\midrule
\textbf{Average} (\emph{macro}) & \highlightpink{54.8} & 59.1 & 56.9 & 54.6 & 57.7 & 61.4 & 72.9 & \gain{18.1} \\
\bottomrule
\end{tabular}
}
\caption{Comparison of baselines with \meta across tasks. Without a Python interpreter, \meta significantly outperforms other methods on the Checkmate-in-One and Sonnet Writing tasks and is on par on most other tasks except Geometric Shapes. \Meta can leverage the Python interpreter in a task-agnostic manner to improve performance significantly across many tasks.}
\vspace{1em}
\label{tab:main-table}
\end{table}

The results of our experiments, summarized in Table~\ref{tab:main-table}, demonstrate the superior effectiveness of our \meta approach compared to the standard zero-shot prompting methods. When we look at the overall performance across all tasks, there is a notable increase in accuracy with \meta, especially when it is augmented with a Python interpreter. Specifically, \meta outperforms standard prompting by 17.1\%, expert (dynamic) prompting by 17.3\%, and multipersona prompting by 15.2\%. Below, we delve into four key insights that emerged from our empirical analysis.

\subsection{Overall Performance}

The \meta approach, particularly when augmented with a Python interpreter, consistently outperforms conventional zero-shot prompting across various tasks. This approach proves to be especially effective in tackling tasks that are heavily reliant on heuristic or iterative trial-and-error problem-solving strategies. 
In the Game of 24 challenge, we see an accuracy improvement of over 60\% compared to the basic standard prompting method (highlighted in pink), about a 15\% gain in Python Programming Puzzles, and close to an 18\% increase in accuracy for Sonnet Writing.
These tasks require complex, iterative, and heuristic search strategies, where conventional single-shot prompting falls short. Conversely, \meta leverages the collective intelligence of various expert personas to iteratively navigate towards a solution, thus fostering a more dynamic and effective problem-solving ecosystem.

Expanding upon its capabilities, \meta appears to be effective in creative writing tasks as well.  In the Shakespearean Sonnet Writing task, for instance, which demands linguistic precision and creative conformity to specific poetic structures, \meta notably enhances performance. While standard prompting methods yield a 62\% accuracy rate, \meta achieves 79.6\% and 77.6\% accuracy, with and without a Python interpreter, respectively. %

In MGSM and Geometric Shapes, the benefits of \meta over the other prompting approaches seem minimal based on the first impression. Nonetheless,\meta does provide 4-6\% gains in Bengali and Telugu, two underrepresented languages with the lowest baseline performances. In Geometric Shapes,\footnote{This task involves naming a shape from its SVG path. Note that the LMs we used did not offer visual capabilities at the time.}
we had expected that GPT-4 to identify the shapes of objects by generating and executing appropriate codes under \meta, but this did not happen. \Meta yielded only a modest 2.4\% gain in this geometric task. We, however, acknowledge that the zero-shot-CoT baseline performed surprisingly better than all the other methods, outperforming \meta with a 10\% accuracy gap. 

While the most significant gains were observed using a Python interpreter, we note that for
the Checkmate-in-One task, \meta achieved a 20.8\% gain even without it.  
Overall, our results highlight the versatility of \meta and underscore its potential for broad application beyond strictly computational problems.

\subsection{Zero-Shot Decomposition, Error Detection, and Aggregation}

The success of our \meta framework lies partly in its strategic use of specialized knowledge, self-collaboration, and implicit verification loops. This approach, as well as multipersona prompting, encourages multi-turn interactions where different personas collaborate to resolve a problem. 

To illustrate how the framework can be beneficial, consider solving multilingual arithmetic problems from the MGSM dataset. GPT-4, under the \meta method, typically follows a three-phase approach: initially translating the problem from the source language (e.g., Bengali) to English, then applying computational expertise (like calling an Expert Mathematician) to find a solution, and finally, conducting an independent or corroborated verification. This unsupervised approach aligns with the multilingual CoT prompting method used by \citet{shi2023language} for MGSM, where the prompt instructs the LM to first translate the problem and then solve it. \Meta performs such translation without explicitly being instructed to do so.

Our structured approach embodies the principle of the wisdom of the crowd~\citep{suzgun-etal-2023-mbrd}, which posits that a collective opinion of a diverse set of critical thinkers often surpasses the insights of individual experts. By harnessing an ensemble of specialized expert models under the guidance of the Meta Model, each contributing from different angles of expertise, we achieve more accurate and reliable problem-solving.

\subsection{Fresh Eyes}

The concept of \textit{fresh eyes} helps mitigate the well-known problem of LMs doubling-down on their mistakes and exhibiting overconfidence \citep[see, e.g.,][]{Zhang2023HowLM}. Fresh eyes are a crucial differentiator between \meta and the multipersona prompting, and thus comparing experimental results demonstrates the advantage. In \meta, fresh perspectives are introduced by engaging experts—or personas—to reassess the problem. This approach provides an opportunity for novel insights and the potential discovery of previously unnoticed incorrect solutions.

Grounded in principles from cognitive psychology, fresh perspectives can lead to more creative problem-solving and error detection. When individuals or models approach a problem without preconceived notions, they are more likely to consider alternative solutions and identify errors that might have been overlooked. Fresh eyes may help avoid cognitive biases such as anchoring, confirmation bias, as well as overconfidence.

Consider the following summary of an execution, which illustrates the benefit of ``fresh eyes'' in practice. Say the task is the 24 game, e.g., to use each of the numbers 6, 11, 12, and 13, exactly once, in an arithmetic expression whose value is 24. The history may look something like the following:

\begin{enumerate} 
    \item The Meta Model proposes consulting experts in mathematics, problem-solving, and Python programming. It emphasizes the need for accuracy and adherence to constraints, suggesting the involvement of another expert for review if needed.
    \item An expert proposes a solution, which a second expert identifies to be incorrect, and the Meta Model suggests writing a Python program to find a valid solution.
    \item A programming expert is consulted to write a program.
    \item Another programming expert identifies an error in the script, modifies it, and then executes the revised script.
    \item A mathematics expert is consulted to verify the solution output by the program.
    \item After this verification, the  Meta Model outputs it as the final answer.
\end{enumerate}

This example underscores how \meta, incorporating fresh perspectives at each step (since the expert's prompt does not include the whole history), not only finds solutions but also effectively identifies and corrects errors. The diversity of perspectives, ranging from problem-solving strategies to technical execution and verification, demonstrates how different angles of expertise contribute to a more robust and reliable problem-solving process.

\subsection{Real-Time Code Execution}
The introduction of a Python expert for code generation and execution within our \meta framework leads to significant advancement in tackling algorithmic challenges. This enhancement is evident in Python Programming Puzzles, where the integration of the Expert Python into the \meta framework elevates the success rate from 32.7\% to 45.8\%. This improvement primarily arises from the Meta Model's ability to use a Python expert for generating and executing code based on natural-language instructions. Real-time code execution enables instant validation and optimization of solutions, substantially improving both the efficiency and precision of problem-solving. 

This enhancement is not confined to a single task type, however. In tasks such as the Game of 24 and Word Sorting, accuracy rates increase by 56.0\% and 15.6\%, respectively, with the integration of a Python interpreter into \meta. (When compared with the baseline standard prompting, the accuracy gains correspond to 64.0\% and 19.2\%, respectively.) These improvements highlight the significant role of code generation and execution in enhancing the effectiveness of the \meta framework, demonstrating its transformative impact across various computational tasks. Overall, integrating a Python interpreter results in an average performance improvement of an additional 11.5\% across different tasks compared to \meta without a Python interpreter. 

However, the introduction of real-time code execution also brings essential security considerations. Establishing such a system requires a secure and controlled environment to mitigate risks such as data breaches and system vulnerabilities. Therefore, the deployment of a Python interpreter within the \meta framework should be fortified with a secure sandbox. These measures are crucial to ensure the system's integrity and the protection of user data, ensuring that the advantages of improved problem-solving efficiency are not compromised in any way by security and privacy concerns, among other issues.

\begin{figure*}[!ht]
    \centering
    \includegraphics[width=0.99\linewidth]{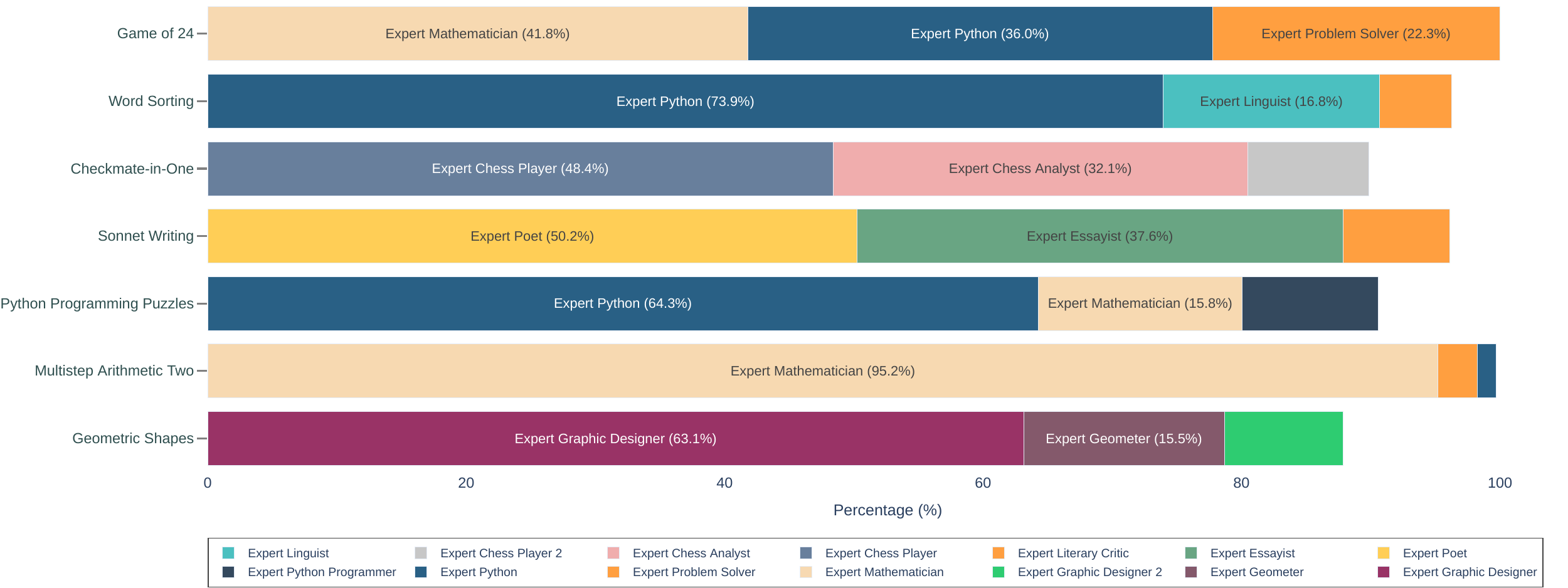}
    \caption{Distribution of experts conjured by the Meta Model in experiments \emph{involving} a Python interpreter. The remaining blank space represents a combination of experts that were employed infrequently.
    }
    \label{tab:expert-distribution-per-task}
\end{figure*}

\begin{figure*}[!ht]
    \vspace{0.2em}
    \centering
    \includegraphics[width=0.99\linewidth]{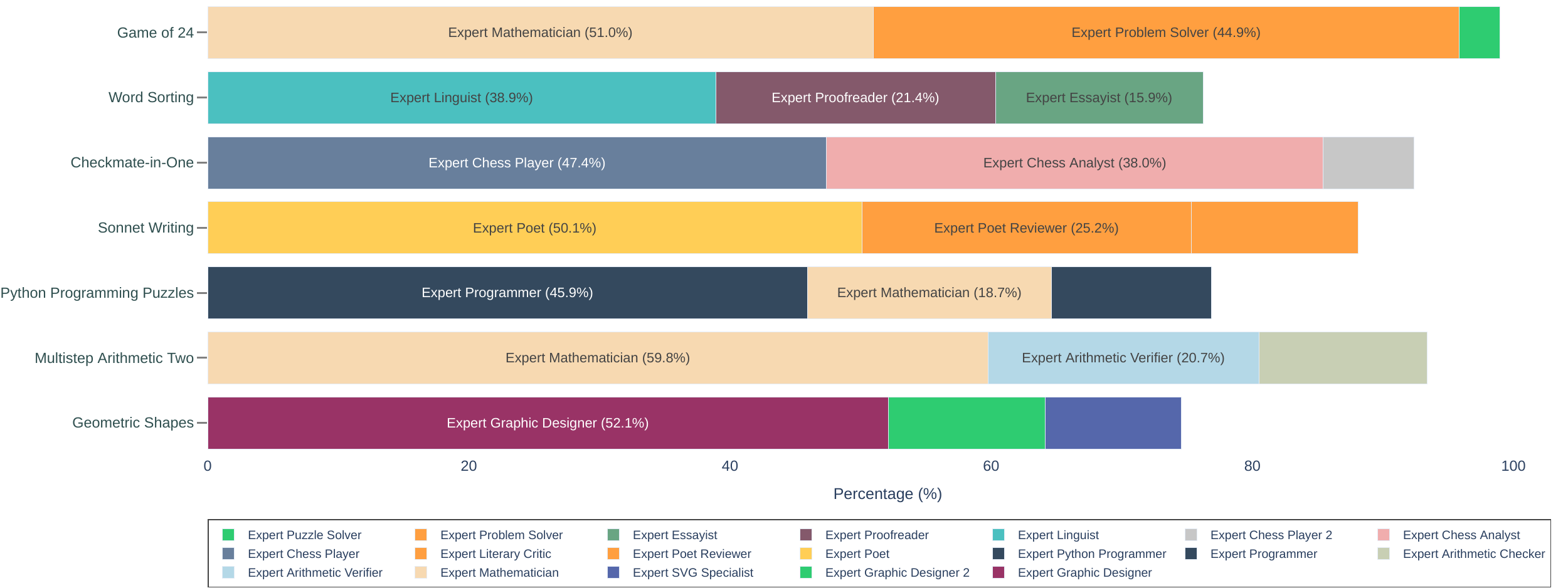}
    \caption{Distribution of experts conjured by the Meta Model in experiments \textit{without} the use of a Python interpreter.
    }
    \label{tab:expert-distribution-per-task}
\end{figure*}

\section{Further Discussion}

\subsection{Additional Analysis of Meta Prompting}

\textbf{Analysis of Expert Types Used in Meta Prompting.} The Meta Model's dynamic selection of expert types distinctly illustrates its adaptability and strategic alignment with specific task requirements. Analyzing tasks with and without a Python interpreter offers insightful contrasts in the model's expert choices, influenced by the available tools and task characteristics. In scenarios where a Python expert is explicitly mentioned for code generation and execution, there is a noticeable preference for technical and computational expertise. For example, in Python Programming Puzzles, the Meta Model frequently utilizes Expert Python, Expert Mathematician, and several tiers of Expert Python Programmers. This pattern reveals a task-oriented strategy, highlighting a focus on programming and algorithmic problem-solving. Similarly, tasks such as Game of 24 and Word Sorting prominently feature Expert Python, reinforcing the model's propensity to rely on computational expertise when Python capabilities are accessible.

In contrast, for \meta without a specific Python expert, the spectrum of experts employed is more diverse. Tasks like Geometric Shapes predominantly involve design and geometry experts (e.g., Expert Graphic Designer and Expert Geometer), indicating a pivot towards visual and spatial problem-solving rather than computational approaches. This task illustrates where the Meta Model may have made a poor choice of experts, and in particular it might have been more preferable to use an expert in SVG visualizations. In Sonnet Writing, the Meta Model naturally leans on literary experts, notably Expert Poet and Expert Literary Critic, emphasizing creative and linguistic skills. This pattern demonstrates the Meta Model's ability to dynamically tailor its expert engagement to the demands of the task, utilizing technical experts for computational challenges and a varied range of non-computational expertise for creative or abstract tasks.

\textbf{Number of Rounds Taken to Reach a Solution.}
Examining the \meta experiments involving a Python expert reveals that the average number of rounds required to reach a solution in the Meta Model varies significantly across tasks, indicative of their complexity and specific nature. Simpler tasks, such as Word Sorting (3.31 rounds) and Checkmate-in-One (3.48 rounds), typically necessitate fewer rounds, suggesting a more linear and straightforward resolution process, likely due to their clearly defined parameters. Conversely, more algorithmically challenging tasks like Python Programming Puzzles average a higher number of rounds at 6.07, reflecting the nuanced and multifaceted aspects of programming tasks that require extensive interactions for thorough clarification and iterative refinement. The Game of 24 and Multistep Arithmetic Two, with averages around 3.5 rounds, meld computational proficiency with logical reasoning, necessitating additional rounds for accurate and precise solutions. This observed correlation between the number of rounds and the task complexity underscores the Meta Model's proficiency and adaptability. It efficiently manages simpler tasks with minimal interactions while skillfully handling the complexities of more challenging and heuristic-based problems, ensuring precision and efficacy in its solutions. This performance characteristic is particularly critical in environments where efficiency and interaction trade-off are key.

\textbf{Enhancing Solution Reliability through Systematic Verification.} The Meta Model's systematic verification protocol strengthens the reliability and robustness of its solutions. Fundamental to this approach is the consistent practice of consulting an expert for validation before finalizing responses, a principle applied across diverse tasks. This method is further evidenced by the detailed interaction data. In tasks such as Checkmate in One, for instance, the Meta Model employs a two-step verification strategy. Initially, it consults an Expert Chess Player to come up with a solution, followed by a critical verification from an Expert Chess Analyst, ensuring strategic correctness. A similar approach is adopted in Sonnet Writing too, where an Expert Poet drafts the sonnet, and an Expert Poet Reviewer or Expert Essayist reviews it, making sure that the solution adheres to the strict rhyme scheme. This unsupervised but rigorous verification process extends to complex tasks like Game of 24 and MGSM, involving both external expert consultations and internal reviews. By integrating this dual verification mechanism, the model significantly enhances solution accuracy and reliability, essential for real-world applications where precision is paramount.

\textbf{Navigating No-Solution Territories.}
\Meta enables the Meta Model to acknowledge the absence or impossibility of a valid solution or its inability to find one more frequently than other prompting methods.
In 100 examples of the Game of 24, the model reports no solution 9 times with Expert Python and 15 times without it, compared to the mere 2 instances under standard prompting. In Checkmate, across 250 examples, it admits to no solution 12 times without Expert Python and 10 times with it, a rarity in multipersona and standard prompting. While there were always solutions, it is arguably preferable to abstain from answering rather than provide an incorrect answer. Typically expressed as ``No valid solution found'' or more explicitly as ``There is no solution to the 24 game with these numbers given the constraints,'' these acknowledgments are likely the result of the model's verification and feedback loop, emphasizing accuracy and confidence over speculative but incorrect responses.

\textbf{Setting the Bar High: GPT-4's Zero-Shot Task Solving Capabilities.} 
Even without the enhanced capabilities of \meta, GPT-4 stands out as an effective zero-shot task solver under standard prompting conditions. Its performance across various tasks, including Python Programming Puzzles and MGSM, is remarkable, particularly when compared to other LMs as highlighted by \citet{GPT4}. %
GPT-4 excels as a task-agnostic solver, %
capable of processing and responding to diverse queries effectively. A significant attribute of GPT-4 is its proficiency in following instructions. Given clear and unambiguous natural-language instructions, the model demonstrates a high level of compliance and accuracy. This aspect of instruction-following is also a cornerstone of our \meta framework, where we leverage GPT-4’s capabilities. Our experiments reinforce that GPT-4 excels in code generation, demonstrates impressive zero-shot reasoning, and engages effectively in role-playing, solidifying its position as a versatile and reliable LM.

\textbf{Limited Performance Improvement with GPT-3.5.}
In comparison to GPT-4, GPT-3.5 demonstrates a more limited scope of performance enhancement across various tasks. Although it shows notable improvements in specific tasks such as Sonnet Writing and Checkmate-in-One, its capabilities do not consistently surpass baseline standards or zero-shot CoT prompting methods in other tasks, notably Word Sorting and Multiple Arithmetic Two. Our qualitative analysis suggests that GPT-3.5 may not be as effective as GPT-4 in simulating role-playing scenarios or managing extended context windows. This observation leads us to believe that factors such as the scale of the model, the quality and size of the instruction-following corpus may be significantly influencing the efficacy of the \meta approach. Furthermore, it appears that the advantages offered by \meta may even emerge more prominently at larger model scales.

\subsection{Limitations and Failure Modes of Meta Prompting}

The \meta framework, despite its innovative approach, encounters several notable limitations, including cost efficiency, scalability, operational linearity, domain restrictions, information transfer challenges, and response patterns. A primary limitation is the elevated cost associated with multiple model calls. In our setup using GPT-4, the dual role of the Meta Model and the experts, distinguished by unique instructions, incurs substantial costs under the GPT-4 API pricing model. This cost factor diminishes the effectiveness of \meta in smaller models like ChatGPT, which lack the comprehensive capabilities of GPT-4. Consequently, \meta, though insightful, can become prohibitively expensive due to extensive model interactions and lengthy message histories. However, these costs will decrease as the costs of LMs decrease. Note that recent OpenAI API features announced after the experiments were run, namely the ability to run code in a sandbox directly through the API, could significantly decrease the costs of our system.

Another critical limitation is the requirement for substantial scale and a considerable context window. GPT-4 fits this criterion, but smaller models such as ChatGPT fall short. \Meta's design, characterized by extensive message histories, demands an LM capable of handling and retaining lengthy textual information, a feature not universally present in all LMs.
Operational efficiency is also challenged by the linear (sequential) nature of \meta. The framework, in its current form, processes steps one at a time, relying on the outcome of preceding calls. This dependency constrains the possibility of parallel processing, impacting the speed and efficiency of the system.

Additionally, our research confined \meta within a closed-domain system. Nevertheless, the framework's potential extends to incorporating external resources such as APIs, specialized finetuned models, search engines, or computational tools. More expansive implementations like AutoAgents~\citep{chen2023autoagents} and AutoGen~\citep{wu2023autogen}, which include higher-level planning and diverse cooperation mechanisms, offer a glimpse into future directions. In subsequent versions, the Meta Model could benefit from refining or summarizing its history before advancing, optimizing the relevance and efficiency of the process. There is also untapped potential in concurrently summoning multiple experts or utilizing a single expert with varied temperature parameters to synthesize their outputs.

A practical challenge faced is the Meta Model's occasional oversight in conveying necessary information to experts, forgetting that experts can only access data adhering to a certain format (within triple quotes in our system). This oversight can lead to unintended confusion and underscores the need for improved information management. Lastly, the Meta Model's response pattern, particularly in tasks with lower performance, often includes apologies, such as “Apologies for the confusion in my previous response” or “I apologize for the previous incorrect solution.” This behavior likely stems from its training on instruction-following data.

\section{Related Work}
This section seeks to contextualize our proposed \meta approach amidst recent advancements in prompting strategies and scaffolding techniques. We provide a brief overview of these developments, highlighting their relevance and connections to our work.

\textbf{Enhancing Reasoning in Language Models through Prompting.} 
Recent efforts in LM scaffolding and prompting methods have significantly boosted the arithmetic and commonsense reasoning capabilities of LMs. The chain-of-thought (CoT) prompting \citep{wei2022chain} and its variants---including least-to-most \citep{zhou2023leasttomost}, zero-shot CoT \citep{ZeroShotCoT}, self-ask~\citep{press2022selfask}, ask-me-anything~\citep{arora2023askmeanything}, decomposed prompting~\citep{khot2023decomposed}, and auto-CoT~\citep{zhang2023automatic}---have marked a paradigm shift in how LMs process complex queries. These methods encourage LMs to adopt human-like, sequential thinking processes, breaking down intricate questions into simpler subtasks and systematically solving them before presenting a final answer. 
Multiple studies \citep[][\textit{inter alia}]{wei2022emergent, madaan2022text, shi2023language, drozdov2023compositional,fu2023complexitybased, suzgun2023bbh} have shown the efficacy of these prompting methods across a broad set of tasks and benchmarks. More recent innovations such Tree-of-Thought~\citep{yao2023tree}, Graph-of-Thought~\citep{besta2023graph}, Program-of-Thought~\citep{chen2023program}, and Skeleton-of-Thought~\citep{ning2023skeleton}, have further enriched this domain; these explore dynamic, non-linear reasoning pathways, broadening the computational and heuristic capabilities of LMs. However, they come with increased resource demands and greater time complexity, require multiple manual prompt crafting, and are often specialized for particular types of tasks.

\textbf{Iterative Self-Feedback and Refinement Mechanisms.} 
Recent instruction-following techniques and data-collection efforts have expanded the capabilities of LMs to follow instructions, emulate certain aspects of human behavior, and assist in tasks such as annotation and evaluation~\citep{haluptzok2022language,aher2023llmssimulation}. LMs such as GPT-4, PaLM, and Llama are now capable of effectively integrating self-feedback and refinement mechanisms through prompting and can leverage their own natural-language outputs to guide their behaviour and improve decision-making. SayCan~\citep{ahn2022saycan} and 
Inner Monologue~\citep{huang2023inner} are early examples showcasing the benefits of inner dialogues in a closed-loop system for robotic control and action planning. Reflexion~\citep{shinn2023reflexion} builds upon these studies and focuses on natural-language generation and reasoning tasks. It functions as a policy optimization mechanism through natural language feedback, using self-feedback and self-reflection to influence and correct behaviors in LMs, and has shown considerable success in preliminary experiments. In a more innovative vein, the Self-Taught Reasoner approach~\citep[STaR;][]{zelikman2022star} iteratively trains an LM on its own outputs to refine initial rationales for more accurate solutions, leading to enhanced reasoning skills. Other notable methods such as 
Critic~\citep{gou2023critic}, 
Iterative Refinement~\citep{chen2023iterative},
RCI~\citep{kim2023language},
Re$^{3}$~\citep{yang-etal-2022-re3},
Refiner~\citep{paul2023refiner},
Self-Critique~\citep{saunders2022selfcritique}, 
Self-Correction~\citep{welleck2023selfcorrection}, 
Self-Eval, 
Self-Debug~\citep{chen2023teaching}, 
Self-Edit~\citep{zhang2023selfedit}, 
Self-Evolve~\citep{jiang2023selfevolve},
Self-Taught Optimizer~\citep[SToP;][]{zelikman2023stop},
and so forth,
illustrate how verbal feedback, both internal and external, can significantly improve the accuracy, quality, and robustness of model outputs across various tasks and setups.

\textbf{Exploring Role-Playing in Language Models.}
The integration of role-playing and self-collaboration concepts into LMs, grounded in cognitive psychology and developmental education principles, has emerged as a useful method for augmenting LMs' problem-solving capabilities and optimizing their internal domain-specific knowledge and expertise. Recent  studies~\citep{park2022social,Park2023GenerativeAgents,li2023camel, xu2023expertprompting, fu2023improving, deshpande2023toxicity} have shown that endowing instruction-following LMs with ``expert'' personas or roles enhances the quality and accuracy of their output. In particular,  approaches like CAMEL~\citep{li2023camel} and Expert Prompting~\citep{xu2023expertprompting}, which involve dynamically assigning personas to a single LM, have been shown to yield higher quality and more reliable responses than models without designated personas. 
Further investigations~\citep{chen2023autoagents,chen2023agentverse,du2023multiagent,hao2023chatllm,liang2023encouraging,liu2023dynamic,jiang2023llm,xiong2023diving,zhang2023cumulative} demonstrate that assigning multiple expert identities or roles to a single LM, tailored to specific tasks or problems, and prompting it to conduct multi-round internal dialogues---similar to a team of experts discussing and refining ideas---amplifies the reliability and comprehensiveness of the LM's analysis; this leads to more well-rounded and thorough solutions. These studies advocate a complementary approach wherein multiple instances of an LM propose, debate, and refine their individual responses and reasoning in successive rounds, culminating in a unified final answer. This role-playing concept has shown to significantly improve mathematical and strategic reasoning across various tasks. Moreover, it improves the factual accuracy of the generated content, thereby reducing erroneous or fabricated responses.

\textbf{Autonomous Decision-Making and Execution in Multi-Agent LM Systems.}
There has been a growing interest in using LMs for autonomous decision-making and task execution. Open-source projects like
\hyperlink{https://github.com/Significant-Gravitas/AutoGPT}{Auto-GPT}, 
\hyperlink{https://github.com/reworkd/AgentGPT}{Agent-GPT},
\hyperlink{https://github.com/yoheinakajima/babyagi}{Baby-AGI}, and
\hyperlink{https://github.com/langchain-ai/langchain}{LangChain}
are notable efforts developing agent protocols that are capable of planning, decision-making, and executing tasks end-to-end, with minimal or no human intervention. These systems highlight the potential and risks of LMs, which go beyond performing predefined tasks to adapting, learning, and autonomously executing decisions in real time. As discussed by \citet{Masa2023AutoGPT}, those autonomous models might be exploited by individuals with malicious intents and pose threats to humanity. There is also the dilemma of accountability: who bears responsibility when an LM-driven autonomous agent produces an inappropriate or criminal action? Ensuring safety and security with these agents is crucial, given its potential for mishaps or exploitation by malicious actors, and its vulnerability to cyber-attacks.

\textbf{Integration of External Tools and APIs into Language Models.} 
As LMs continue to evolve, the integration of external tools is becoming increasingly important. This tool-use integration, often achieved through in-context learning~\citep[e.g.,][]{cai2023large} or finetuning~\citep[e.g.,][]{Schick2023ToolformerLM}, allows LMs to effectively engage with real-world scenarios and tackle a diverse range of dynamic tasks. Recent advancements~\citep{
cai2023large,
gao2023assistgpt,
gou2023critic,
hao2023toolkengpt,
khattab2023dspy,
lu2023chameleon,
qiao2023making,
paranjape2023art,
patil2023gorilla,
Schick2023ToolformerLM,
yang2023gpt4tools,
yuan2023craft} have enabled LMs to perform accurate calculations, retrieve up-to-date information from search engines or databases, and interact with APIs, making them crucial for complex, multimodal real-world problems. 
OpenAI's incorporation of predefined APIs and plugins into ChatGPT underscores the importance of external integration in developing a comprehensive LM ecosystem. However, most approaches often limit themselves to a select group of tools or domain-specific resources, posing challenges in adapting to new domains~\citep{lu2023chameleon}. Our \meta approach, as detailed in Section~\ref{sec:meta-promting}, treats the LM as an independent tool and expert, available on-demand for specific tasks. Furthermore, incorporating a Python interpreter---through Expert Python---to execute and evaluate model-generated code has been instrumental in enhancing both accuracy and efficiency in various tasks.

\section{Conclusion}

In this work, we have introduced and examined \meta, a simple yet powerful scaffolding technique that enhances the performance of language models in a task-agnostic manner. This approach leverages a language model to act as both a central conductor and a group of expert instances, thereby endowing traditional models with dynamic, multi-functional capabilities. A noteworthy aspect of \meta lies in its proficiency to decompose complex tasks, engage distinct expertise for each component, and then integrate the varied outputs seamlessly. Demonstrating significant, double-digit improvements across a series of tasks, ranging from challenging arithmetic puzzles like the Game of 24 to the creative literary exercise of Shakespearean Sonnet Writing, \meta promises to grow more potent and cost-efficient as language models continue to evolve, offering exciting prospects for future applications.

\section*{Acknowledgements}
We would like to thank
Federico Bianchi,
Annabelle Carrell,
Tayfun G\"{u}r,
Dan Jurafsky,
Suproteem Sarkar,
Scott Duke Kominers,
Lester Mackey,
Neil Mallinar,
\c{S}ule Kahraman,
Deniz Kele\c{s},
Luke Melas-Kyriazi,
Drew Pendergrass,
Faiz Surani,
Garrett Tanzer,
Michael Wornow, and
Eric Zelikman
for their valuable comments, useful suggestions, and support.

\clearpage

\bibliography{refs}
\bibliographystyle{ACM-Reference-Format}

\clearpage
\appendix

\end{document}